\documentclass{article}
\usepackage{kr}

\usepackage{times}
\usepackage{soul}
\usepackage{url}
\usepackage[hidelinks]{hyperref}
\usepackage[utf8]{inputenc}
\usepackage[small]{caption}
\usepackage{graphicx}
\usepackage{amsmath}
\usepackage{amsthm}
\usepackage{booktabs}
\usepackage{algorithm}
\usepackage{algorithmic}

\usepackage{subcaption}
\usepackage[nolist]{acronym}
\usepackage{listings}
\usepackage{xcolor}
\usepackage{float}
\usepackage{todonotes}
\usepackage{comment}

\definecolor{lbcolor}{rgb}{0.95,0.95,0.95}  
\acrodef{vr}[VR]{virtual reality}
\acrodef{neem}[NEEM]{narrative-enabled episodic memory}
\acrodefplural{neem}[NEEMs]{narrative-enabled episodic memories}
\acrodef{krr}[KR\&R]{knowledge representation and reasoning}
\acrodef{soma}[SOMA]{Socio-Physical Model of Activities}
\acrodef{cpl}[CPL]{\textsc{Cram} planning language} 
\acrodef{tamp}[TAMP]{Task and Motion Planning}
\acrodef{robcog}[RobCoG]{Robot Common-Sense Games}
\acrodef{llm}[LLM]{Large Language Model}

 \newcommand\copyrighttext{%
 \footnotesize \textcopyright 2023 KR Inc. Personal use of this material is permitted.
 Permission from KR Inc. must be obtained for all other uses, in any current or future 
 media, including reprinting/republishing this material for advertising or promotional 
 purposes, creating new collective works, for resale or redistribution to servers or 
 lists, or reuse of any copyrighted component of this work in other works.}
\newcommand\copyrightnotice{%
\begin{tikzpicture}[remember picture,overlay]
\node[anchor=south,yshift=10pt] at (current page.south) {\fbox{\parbox{\dimexpr\textwidth-\fboxsep-\fboxrule\relax}{\copyrighttext}}};
\end{tikzpicture}%
}

\title{Knowledge-Driven Robot Program Synthesis from Human VR Demonstrations}
\author{%
Benjamin Alt$^{1,2}$\and
Franklin Kenghagho Kenfack$^2$\and
Andrei Haidu$^2$\and
Darko Katic$^1$\and
Rainer Jäkel$^1$\and
Michael Beetz$^2$\\
\affiliations
$^1$ArtiMinds Robotics, Karlsruhe, Germany\\
$^2$Institute for Artificial Intelligence, University of Bremen, Germany\\
\emails
\{benjamin.alt, darko.katic, rainer.jaekel\}@artiminds.com,
\{fkenghag, haidu, michael.beetz\}@uni-bremen.de
}

\begin{document}

\maketitle

\copyrightnotice

\global\csname @topnum\endcsname 0
\global\csname @botnum\endcsname 0

\begin{abstract}
Aging societies, labor shortages and increasing wage costs call for assistance robots capable of autonomously performing a wide array of real-world tasks. Such open-ended robotic manipulation requires not only powerful knowledge representations and reasoning (KR\&R) algorithms, but also methods for humans to instruct robots what tasks to perform and how to perform them. In this paper, we present a system for automatically generating executable robot control programs from human task demonstrations in virtual reality (VR). We leverage common-sense knowledge and game engine-based physics to semantically interpret human VR demonstrations, as well as an expressive and general task representation and automatic path planning and code generation, embedded into a state-of-the-art cognitive architecture. We demonstrate our approach in the context of force-sensitive fetch-and-place for a robotic shopping assistant. The source code is available at \url{https://github.com/ease-crc/vr-program-synthesis}.
\end{abstract}

\begin{figure}
    \centering
    \includegraphics[width=\linewidth]{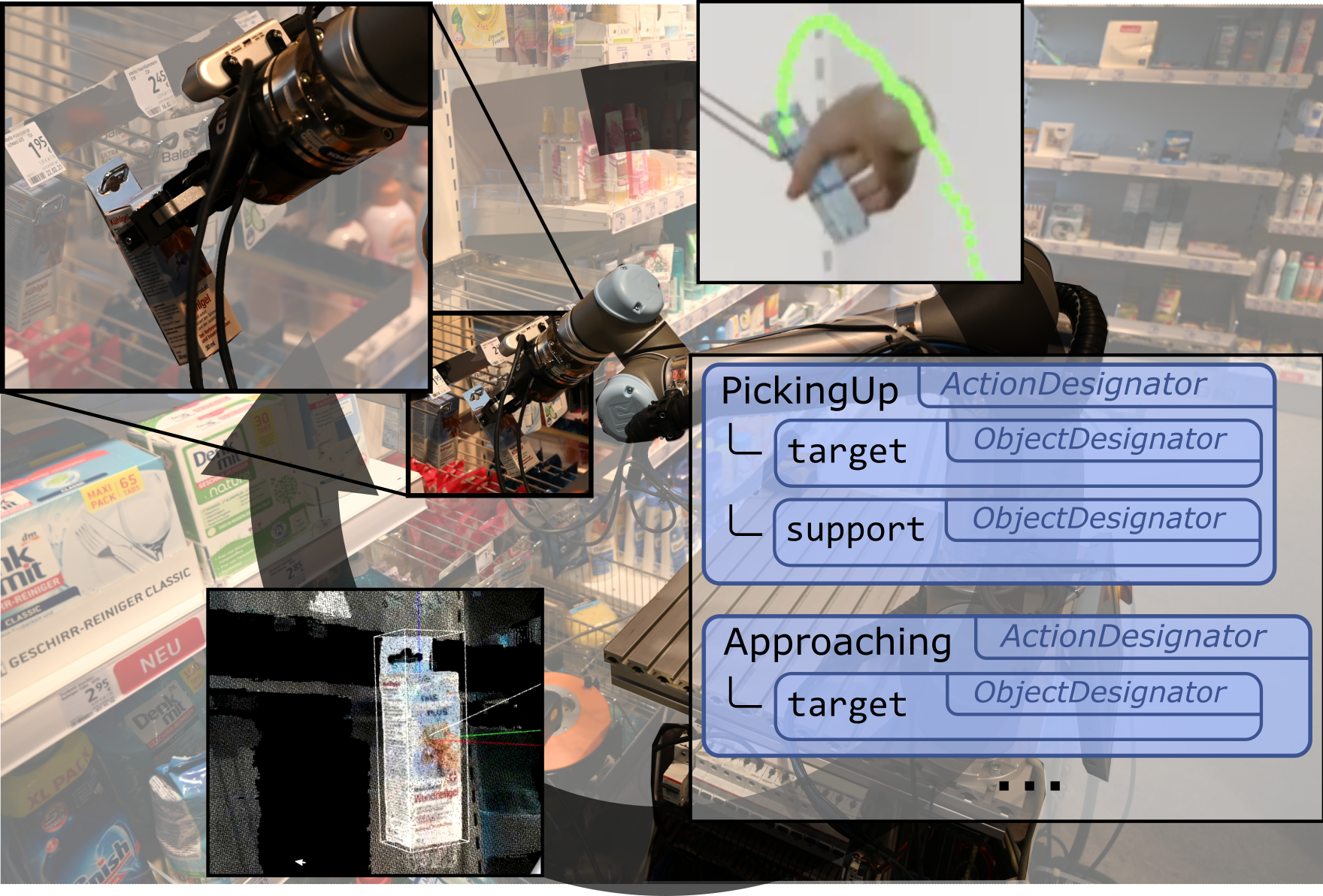}
    \caption{We propose a knowledge-driven approach to convert human demonstrations in virtual reality (top right) to executable robot programs by leveraging semantic task knowledge (center right) and knowledge-augmented perception (bottom).}
    \label{fig:teaser_image}
\end{figure}

\section{Introduction}

Robots are universal manipulators: Their versatility promises a future in which robots assist humans not just on industrial assembly lines, but also in everyday situations such as shopping or elderly care. Many everyday assistance tasks such as setting a table or fetching objects in a supermarket involve open-ended manipulation, requiring the robot to reason about tasks which cannot be exhaustively specified. It would be impossible or prohibitively expensive, for example, to provide a shopping assistance robot with a dedicated control program for each product it may be asked to fetch from a shelf. Instead, the robot should be programmed in a way that is sufficiently general to cover a wide array of objects, while at the same time allowing fine-grained control over important task-specific details such as force limits when handling fragile items like glass or fruit. 

The adoption of robots for everyday assistance tasks will require novel methods of programming. Non-experts should be able to specify tasks, goals and constraints in an intuitive manner. Based on high-level human input, robot control programs should be automatically generated and specialized to meet the requirements of the demonstrated task and environment. Program synthesis for open-ended manipulation tasks requires reasoning about objects and their physical properties as well as adaptability to changes in the environment and the task requirements. ac{krr} approaches permit the design of cognitive systems capable of solving such open-ended tasks in a data- and compute-efficient manner.

\begin{figure*}
    \centering
    \includegraphics[width=\textwidth]{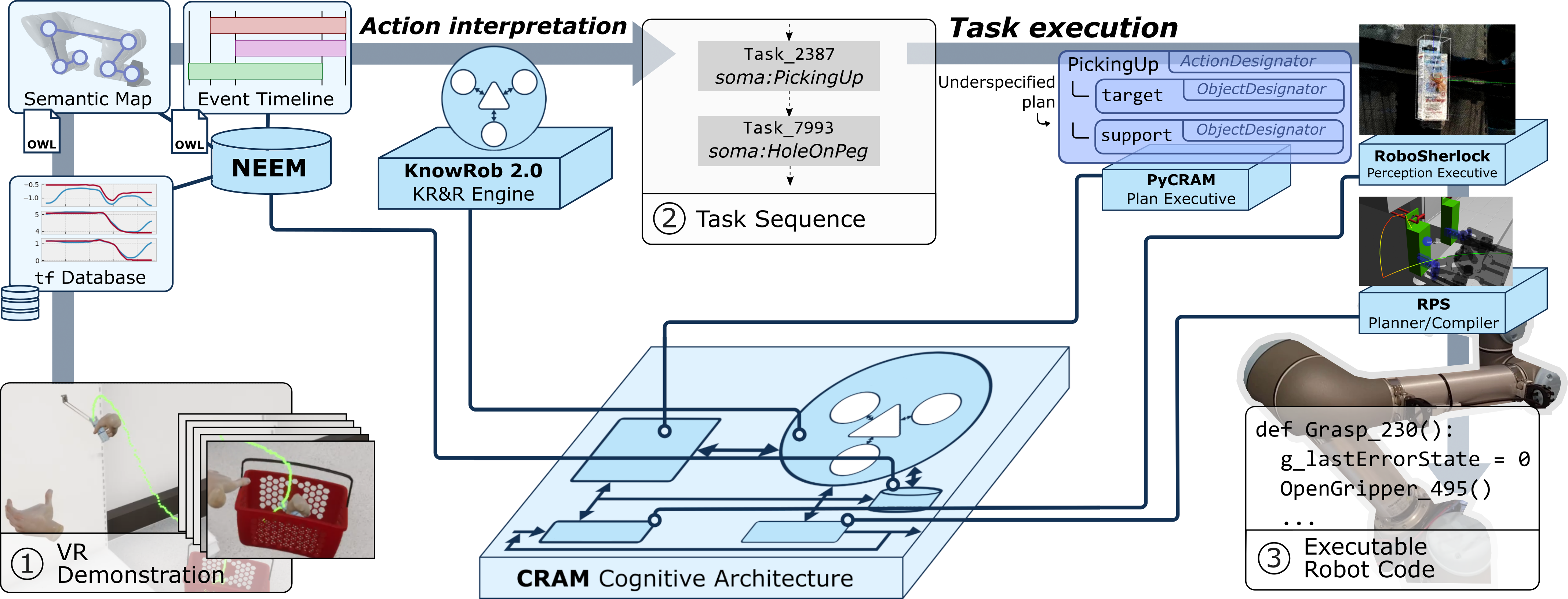}
    \caption[]{Overview of knowledge-driven robot program synthesis from \ac{vr} demonstrations. Humans demonstrate complex manipulation actions in \ac{vr} (1). Using common-sense knowledge, an \ac{krr} engine can interpret this experience data as a sequence of underspecified tasks (2). This sequence is then translated into a plan, which is grounded to the real world via high-level reasoning, knowledge-based perception, motion planning and code generation. The resulting control program can be directly executed on the robot (3).}
    \label{fig:overview}
\end{figure*}

We contribute a system that is capable of inferring executable robot control programs from a single human \ac{vr} demonstration of a manipulation task. We decompose program generation into two steps -- action interpretation and task execution. \textit{Action interpretation} refers to the inference of an underspecified task such as "Hanging an object onto a hanger", which captures the human's intention, from low-level \ac{vr} data. \textit{Task execution}, in turn, generates a fully specified, executable robot control program for the inferred task. By parsing \ac{vr} data into a semantically rich knowledge representation, and by designing and implementing a collection of general ontological knowledge about robot tasks as well as their preconditions and effects, the unification algorithm can be leveraged to infer a robot control program in order to perform the demonstrated tasks. We experimentally validate our system on two challenging real-world assistance tasks in a supermarket environment.
Due to the principled use of knowledge representation and reasoning, our approach generates robust robot programs from a single \ac{vr} demonstration, without requiring training data. Moreover, it immediately generalizes to arbitrary object poses, shapes, and environment configurations. 

\section{Overview}
Fig. \ref{fig:overview} provides an overview of the program synthesis system. Given a human demonstration nof a manipulation task in \ac{vr}, an executable robot program is automatically synthesized via a three-step process:

\subsubsection{Knowledge extraction}
In a first step, the \ac{vr} demonstration is automatically parsed into an episodic memory: A semantically rich representation of human or robot actions which contains both a symbolic log of detected events in terms of an ontology as well as a subsymbolic record of the motions of the agent and objects in the environment (see Sec. \ref{sec:knowledge-representation}). This episodic memory is inserted into the knowledge base of a \ac{krr} engine. Parsing \ac{vr} demonstrations into episodic memories connects the demonstration to the extensive background knowledge provided by upper-level domain and application ontologies.

\subsubsection{Action interpretation}
Episodic memories describe sequences of events, which may be contingent on the agent's (here, the human demonstrator's) abilities and the particular configuration of the \ac{vr} environment. In order to generate executable programs for robots acting in real-world environments, the action sequence is \textit{lifted} from the action (event) to the task (concept) level, discarding event-specific information such as durations or locations and inferring task-relevant information such as roles or constraints (see Sec. \ref{sec:interpretation-of-demonstrated-action-sequences}). This process is not one-to-one: The interpretation of an action can yield a set of possible tasks, depending on the context and the amount of available information. For example, an action during which a lid is moved onto a container can be interpreted as a \texttt{Closing} task, but also as a \texttt{Placing} task. Consequently, the interpretation of an entire demonstration typically yields more than one candidate task sequence.

\subsubsection{Task execution}
Each inferred task sequence can be automatically mapped to a \textit{plan}, which contain a list of actionable steps for the robot to fulfill the tasks (see Sec. \ref{sec:program-generation-and-execution}). These initially underspecified plans are refined into executable robot programs via a flexible reasoning, perception and planning pipeline (see Fig. \ref{fig:program_generation_pipeline}). Inferring underspecified plans enables learning solutions to tasks at a very high level, such as ``placing an object onto a shelf'' or ``taking an object from a hanger'', reducing the amount of required demonstrations to a minimum. The program generation pipeline will produce executable robot code specialized to the current environment, hardware and robot capabilities from these plans.

\section{Knowledge Representation}
\label{sec:knowledge-representation}
Enabling robots to synthesize their own control programs requires reasoning about possible tasks, available actions and the properties of objects in the environment. We leverage methods of explicit knowledge representation to facilitate such reasoning.

\subsection{Ontological Model of Tasks and Events}
\label{sec:ontological-model-of-tasks-and-events}
\begin{figure}
    \centering
    \includegraphics[width=\linewidth]{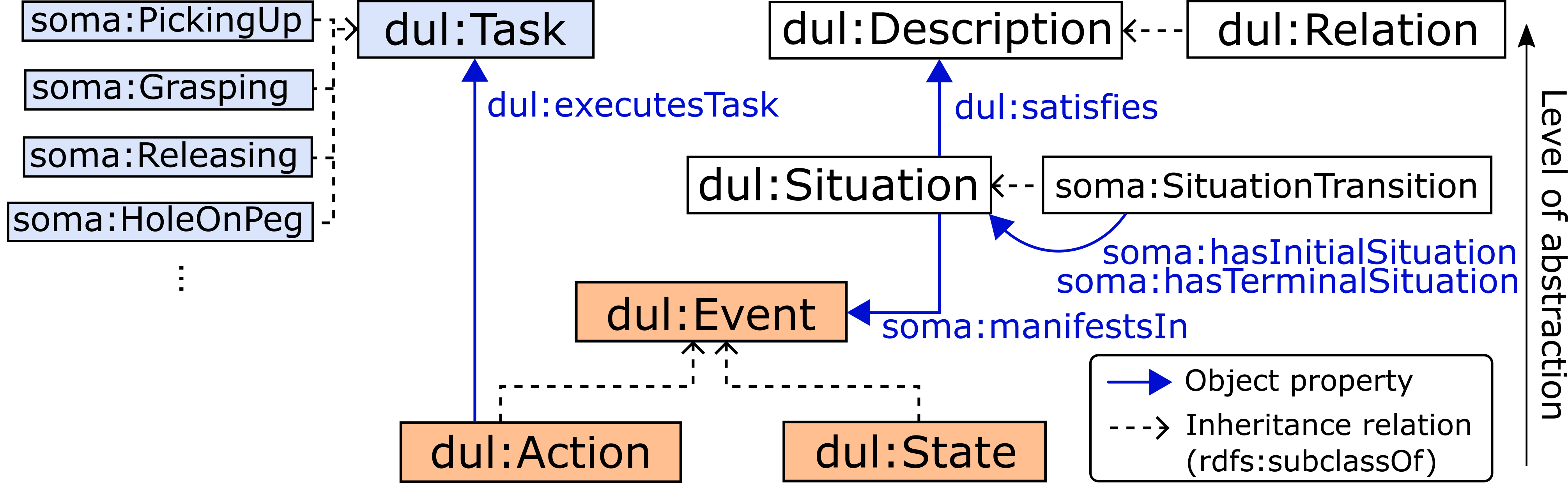}
    \caption{The \ac{soma} model of \texttt{Task}s, \texttt{Event}s and \texttt{Situation}s \protect\cite{besler_foundations_2020}.}
    \label{fig:soma-task-event-situation}
\end{figure}

A knowledge-driven approach to robot program synthesis requires a semantic model of the meaning of tasks, actions and events and their relation to each other. We adopt the model proposed by the SOMA ontology (\aclu{soma}) \cite{besler_foundations_2020}. The goal of robot programs is to cause some \texttt{Event}s\footnote{We use a fixed-width font (e.g. \texttt{Grasping}) to indicate classes, properties etc. as defined in \ac{soma} or other ontologies. For readability, namespace prefixes are omitted or replaced by short-form prefixes (e.g. \texttt{Grasping} or \texttt{soma:Grasping} rather than {\scriptsize \texttt{http://www.ease-crc.org/ont/SOMA.owl\#Grasping}}).} in the world (a robot grasps a bag of chips, lifts it, then puts it into a basket), which bring about one or more \texttt{Situation}s (the bag of chips is in the basket) seen as  consistent with a \texttt{Task} (shopping for chips). In this model, \texttt{Task}s are \texttt{Concept}s, which describe how an \texttt{Event} should be interpreted. In the opposite direction, when a \texttt{Task} is executed, it results in one or multiple \texttt{Event}s (see Fig. \ref{fig:soma-task-event-situation}). Both the interpretation of \texttt{Event}s as \texttt{Task}s and the execution of \texttt{Task}s to produce \texttt{Event}s is highly context-dependent: The robot's gripper making contact with an object can imply \texttt{Grasping} or \texttt{Pushing}, depending on what happens afterwards. Similarly, the task of opening a container can be performed in a variety of different ways, depending on the capabilities of the robot.

Generating executable robot control programs for one or more ambiguously specified high-level tasks is the focus of most prior work on task-level programming. A comprehensive approach to program synthesis, however, must also consider ways for humans to express which tasks the robot should solve (e.g. via natural language or demonstrations in \ac{vr}). Most such modalities require \textit{interpretation}: In the case of natural language, the user's words (descriptions of intended actions) must be interpreted to tasks while taking the larger context into account. ``Smash the button'' may mean destroying it in the context of recycling or vigorously pushing it in the context of machine tending. In the case of \ac{vr} demonstrations, the user takes actions in the \ac{vr} environment, which must be interpreted to tasks depending on the types of objects in \ac{vr}, their properties and the domain: A \ac{vr} demonstration of the same screwing task, for example, will look very different if the demonstrator is using a powered or a manual screwdriver. For these reasons, we consider both the translation of actions into tasks (``interpretation'', see Sec. \ref{sec:interpretation-of-demonstrated-action-sequences}) and the translation of tasks into actions (``execution'', see Sec. \ref{sec:program-generation-and-execution}).

\begin{figure*}
    \centering
    \includegraphics[width=\textwidth]{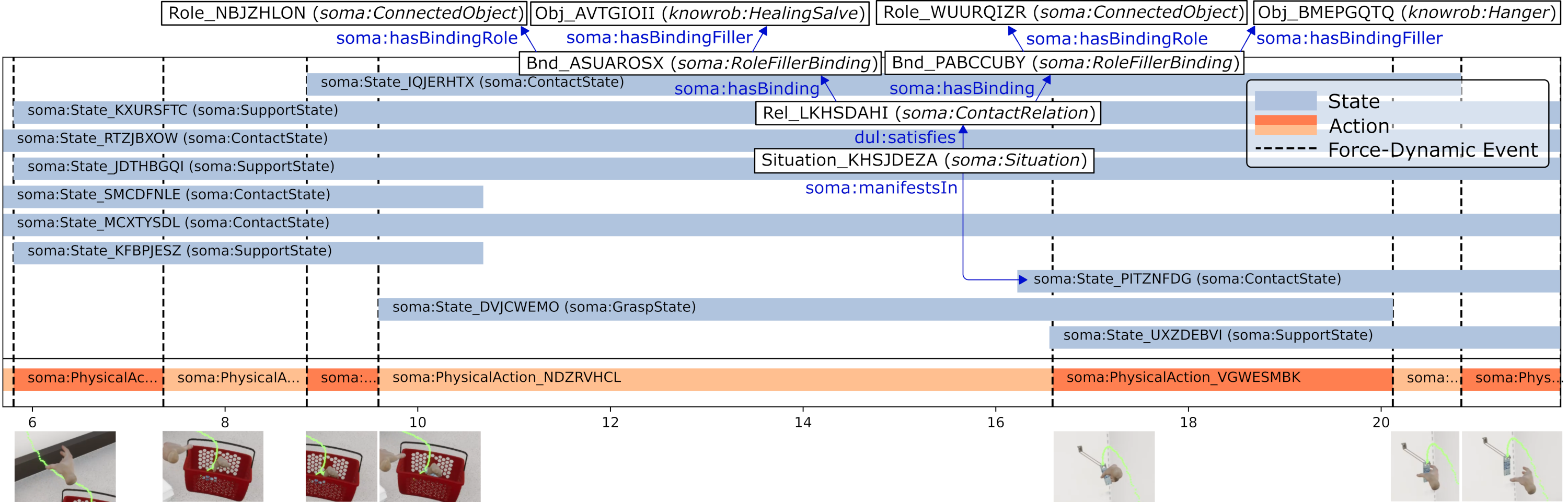}
    \caption{Timeline of a \ac{vr} demonstration segmented by force-dynamic events.}
    \label{fig:event-timeline}
\end{figure*}

\subsection{Representing Demonstrations: Episodic Memories}
\label{sec:representing-demonstrations}
To facilitate reasoning over human demonstrations, we propose to represent them as \acp{neem}, semantically enriched execution traces in the \textsc{Cram} ecosystem \cite{beetz_know_2018}. A \ac{neem} has three components:
\begin{enumerate}
    \item A \textit{semantic map}, which is an ontology (derived from \ac{soma}) containing a description of all relevant agents and objects in the environment, their properties and interrelationships.
    \item An \textit{event timeline}, which is an ontology (derived from \ac{soma}) containing a set of timestamped \texttt{Event} individuals (\texttt{State}s and \texttt{Action}s), as well as semantic annotations of the \texttt{Situation}s manifesting in them (see Fig. \ref{fig:event-timeline}). This realizes the model of tasks and events described in Sec. \ref{sec:ontological-model-of-tasks-and-events}. \texttt{Situation}s are descriptive contexts linking \texttt{Event}s to descriptions of the world which hold over their duration - examples are \texttt{Relation}s between objects (contact, support, containment, ...) or context-dependent object properties (affordances). 
    \item A \textit{database} containing the poses of all relevant agents and objects over time, as well as the motions of their links.
\end{enumerate}
A simplified example for a \ac{neem} is shown in Fig. \ref{fig:event-timeline}. Unlike raw \ac{vr} data, this semantically rich representation of human demonstrations enables robots to reason about demonstrated action sequences as if it was their own experience. Because \ac{neem} experience is expressed in terms of ontologies, differences between the \ac{vr} world and the real setting such as kinematic differences between the human \ac{vr} avatar and the robot or differences in object types or poses can be handled gracefully by the reasoner using ABox reasoning.\footnote{The source code for generating \acp{neem} from \ac{vr} demonstrations is available at \url{https://github.com/ease-crc/vr-neem-converter}.}

\subsection{Representing Task Knowledge}
\label{sec:representing-task-knowledge}
Our approach to program synthesis hinges on the interpretation of demonstrated actions as abstract tasks. In addition to a semantically rich representation of the demonstrations, this requires a similarly rich representation of task types and their meaning. In natural language, a task such as \texttt{Reaching} can be described as ``moving the hand toward an object''. A structured knowledge representation enable robots to perform common-sense reasoning about what tasks are feasible (``a robot can only grasp if its gripper is empty'') or what makes a task successful (``after placing an object, the object is supported by a surface''). This knowledge representation must be sufficiently universal to cover arbitrary domains, and avoid requiring excessively specific demonstrations or descriptions, but it must also allow to draw specific inferences about what the robot must do to achieve a task in a given environment. We propose to describe tasks using a hybrid representation in which the task taxonomy is defined in terms of an ontology, while task semantics are defined as a set of Prolog rules over the ontology.

\subsubsection{Task ontology}
To define what a task is, which classes of tasks exist (e.g. \texttt{Placing}, \texttt{Grasping}, \texttt{Opening}), which inheritance relationships exist between the different task types (e.g. \texttt{Grasping rdfs:subclassOf Manipulating)} and how this task hierarchy relates to other concepts such as actions or parameters, we use and extend the \ac{soma} ontology. A \texttt{Task} is an \texttt{EventType} \textit{classifying} an \texttt{Event}. Defining tasks in terms of \ac{soma} ensures compatibility with other tools and frameworks in the \textsc{Cram} cognitive architecture. When new domain- or application-specific tasks are added (e.g. for the experiments in \ref{sec:experiments}), they are declared as subclasses of \texttt{soma:Task}.

\begin{algorithm}[tb]
    \caption{Pre-, runtime and postconditions for a \texttt{PickingUp} task.}
    \label{fig:picking-up-prolog}
    \begin{lstlisting}[language=Prolog]
satisfies_pre(Act, soma:'PickingUp') :-
  % precondition 1: Some object O1 grasped
  has_initial_situation(Act, S1),
  object_grasped(O1, S1),
  % precondition 2: O1 supported by something
  has_initial_situation(Act, S2),
  object_supported(O1, O2, S2).

satisfies_run(Act, soma:'PickingUp').

satisfies_post(Act, soma:'PickingUp') :-
  % postcondition 1: O1 still grasped
  has_terminal_situation(Act, S1)), 
  object_grasped(O1, S1),
  % postcondition 2: O1 not supported anymore
  forall(has_terminal_situation(Act, S2)),
    \+ object_supported(O1, O2, S2)).
    \end{lstlisting}
\end{algorithm}

\subsubsection{Task semantics}
The ontology only specifies which task types are known and how \texttt{Task}s are related to other entities in \ac{soma} such as \texttt{Action}s, \texttt{Situation}s or \texttt{Object}s. We define the task \textit{semantics} (what distinguishes \texttt{Opening} from \texttt{Closing}? How are observed actions related to the specific tasks they execute?) as a set of Prolog rules in the \textsc{KnowRob} \ac{krr} engine (see Fig. \ref{fig:picking-up-prolog} for an example). This set of Prolog rules can be understood as axiomatic knowledge about tasks and how they relate to (observed) actions. We follow the established method of defining different tasks in terms of their preconditions, runtime conditions and postconditions. For a given task type TT and action A, a task of type TT is executed by A if
\begin{enumerate}
    \item the state of the world just before A begins are consistent with the preconditions of TT,
    \item the state of the world during A is consistent with the runtime conditions of TT,
    \item the state of the world just after A is finished is consistent with the postconditions of TT.
\end{enumerate}
In principle, tasks can be defined in terms of pre-, runtime and postconditions directly in the ontology via TBox axioms or SWRL rules. This would make it very easy to test if an observed action can be interpreted as executing a given task (because then, the ontology would remain consistent). But it makes it very hard to infer the set of tasks which an action possibly executes: Due to the open world assumption, the pre-, runtime and postconditions of some task would be satisfied unless something about the action violates them, and the set of preconditions must cover the infinitely large set of states of the world which preclude the task. By contrast, in a closed-world reasoning system like Prolog, the pre-, runtime and postconditions of tasks are by default violated. In such a system, defining the preconditions of a task only requires predicates covering the much smaller set of states of the world in which the task \textit{can} possibly be executed.

An additional advantage of using a hybrid representation is is that tasks can be specified in very abstract and concise terms. To specify e.g. the preconditions of \texttt{PickingUp}, it suffices to state that some object must currently be grasped, and that this object is supported by some other object (see Fig. \ref{fig:picking-up-prolog}). The reasoning system can infer the proper interpretation of ``being grasped'' or ``being supported'' depending on the capabilities of the agent and the properties of the involved objects.

\section{Interpretation of Demonstrated Action Sequences}
\label{sec:interpretation-of-demonstrated-action-sequences}

In order for a robot to automatically generate its own control programs from tasks, it must first \textit{interpret} the demonstrated human actions as tasks. Informally, action interpretation aims to uncover the intent behind a demonstration: ``What did the human want to demonstrate?'' In our experiments, we consider human \ac{vr} demonstrations represented as \acp{neem}. In principle, real-world human demonstrations or even experience data from other robots can also be used as inputs for program synthesis.

The hybrid task representation outlined in Sec. \ref{sec:representing-task-knowledge} permits the automatic interpretation of action sequences to high-level tasks. Because the task definition comprises a set of Prolog predicates stating a task's preconditions, runtime conditions and effects, Prolog's unification algorithm can be used to generate the set of tasks an action can be interpreted to execute, using the following simple inference rule:

\begin{algorithm}
\caption{Semantic interpretation of an \texttt{Action}}
\label{fig:interprets-to-prolog}
\begin{lstlisting}[language=Prolog]
interprets_to(Act, Tsk) :-
  has_task_type(Tsk, TskType),
  satisfies_pre(Act, TskType),
  satisfies_run(Act, TskType),
  satisfies_post(Act, TskType),
  parameterize_task(Act, TskType, Tsk).
\end{lstlisting}    
\end{algorithm}

If the pre-, runtime- and postcondition for a \texttt{Task} Tsk of type TskType are satisfied, \texttt{parameterize\_task} will instantiate a \texttt{Task} individual for the given \texttt{Action} Act. Note that \texttt{interprets\_to} can be queried repeatedly for the same Act, and will yield individuals for all possible tasks Act can interpret to. This is to be expected - many actions are inherently ambiguous, and distinguishing between them would require highly specific task types with very narrow task definitions. We leave it to the program generation and execution pipeline to resolve such ambiguities (see \ref{sec:program-generation-and-execution}).

Querying \texttt{interprets\_to} for all \texttt{Action}s in a \ac{neem} yields a \textit{task sequence} (a list of \texttt{Task} individuals). Querying \texttt{interprets\_to} exhaustively will yield all possible task sequences for the \ac{neem}:
\begin{algorithm}
\caption{Semantic interpretation of a \ac{neem}}
\label{fig:interpretation-of-nem}
\begin{lstlisting}[language=Prolog]
actions_interpret_to(ActionSeq, TaskSeq) :-
  findall(Act,
           (member(Act, ActionSeq),
            interprets_to(Act, Tsk)),
         TaskSeq).
\end{lstlisting}
\end{algorithm}

From the resulting task sequences, executable robot programs can be automatically generated (see Sec. \ref{sec:program-generation-and-execution}).\footnote{The source code for the task representation and our reasoner is available at \url{https://github.com/ease-crc/vr-program-synthesis}.}

\begin{figure*}[ht!]
    \centering
    \includegraphics[width=\textwidth]{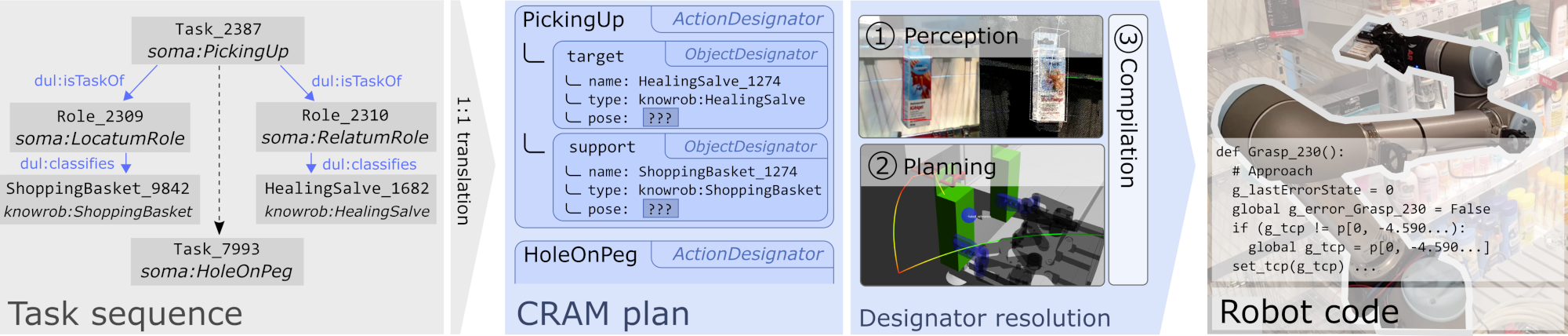}
    \caption{The proposed program generation pipeline. A sequence of \texttt{Task}s is translated into an underspecified plan (here, the real-world poses of the healing salve and shopping basket are not known a priori). Via reasoning, perception and collision-free motion planning, the the plan is ultimately resolved to native robot code.}
    \label{fig:program_generation_pipeline}
\end{figure*}

\section{Program Generation \& Execution}
\label{sec:program-generation-and-execution}

Given a \texttt{Task} individual or a sequence of \texttt{Task} individuals, we propose a program generation and execution pipeline translating underspecified plans to fully specified, executable robot programs. The pipeline combines and extends several elements of the \textsc{Cram} cognitive architecture, such as the CRAM planning language, the \textsc{KnowRob} \ac{krr} engine and the \textsc{RoboSherlock} perception framework.

\subsection{Underspecified Plans}
The \ac{cpl} \cite{beetz_cram_2010} supports hierarchical task-level programming via the concept of \textit{action designators}. A \textit{designator} is a data structure that represents (designates) an entity and associates symbolic and subsymbolic information with it \cite{mcdermott_reactive_1991}. A sequence of (possibly nested) action designators (e.g. \texttt{PickingUp}, \texttt{Transfer}, \texttt{PuttingDown}) forms an \textit{underspecified plan}, where some information required for execution is unknown and must be inferred by a reasoner from a knowledge base (e.g. in which direction a door opens) or determined by perception (e.g. where a target object is located) at runtime. When a \textsc{Cram} plan is executed, all designators are resolved in turn, possibly yielding other designators, until arriving at the lowest-level \textit{motion designators}, which are resolved by executing a planned motion on a robot. We refer to \cite{beetz_cram_2010} for a more detailed overview of \ac{cpl}.

With the task representation proposed in Sec. \ref{sec:representing-task-knowledge}, translating a sequence of \texttt{Task}s into an underspecified plan is straightforward. Because \texttt{Task}s and \ac{cpl} action designators share the same level of abstraction, converting a task sequence to an underspecified plan reduces to instantiating one action designator per \texttt{Task}, as well as object designators or location designators for the paramers of the \texttt{Task}, depending on their \texttt{Role} in the task. A \texttt{PickingUp} task, for example, has two objects associated with it: A \texttt{Locatum} (the primary object, here the picked up object) and a \texttt{Relatum} (the secondary object, here the object supporting or containing the \texttt{Locatum}). It can be directly translated into a \texttt{PickingUp} action designator with a ``target'' property (the \texttt{Locatum}) and a ``support'' property (the \texttt{Relatum}) (see Fig. \ref{fig:program_generation_pipeline}). For this work, we extended PyCRAM\footnote{The Python implementation of \textsc{Cram} (\url{https://github.com/cram2/pycram}).} to support a total of 27 different task types, ranging from prehensile manipulation (\texttt{Grasping}, \texttt{PickingUp}, \texttt{Placing}) to force-controlled insertion and retraction (\texttt{HoleOnPeg}, \texttt{Sliding}, \texttt{Retracting}).

\subsection{Generation of Executable Robot Control Programs}
The generated underspecified plan contains action, object and location designators, which must be grounded (\textit{resolved}) in order for a real robot to perform actions in the real world. 

\subsubsection{Object and location designator grounding}
A \texttt{HoleOnPeg} action designator, for example, encapsulates the series of actions required to thread an object with a hole onto a peg (e.g. hanging a product with a perforated tab onto a hanger in a supermarket). \texttt{HoleOnPeg} is parameterized with two object designators, \textit{hole} (the object to be hung) and \textit{peg} (the hanger). The location of the \textit{hole} is generally not known when the plan is generated, and must be detected at runtime. Moreover, the precise type or instance of the \textit{hole} object may also not be known; if the plan was generated from a \ac{vr} demonstration, the object instance will certainly be unknown (the virtual object will not be present in the real world) and the object type in the demonstration will likely be different from the objects present in the supermarket. The missing properties (pose, type, individual) must be resolved using contextual knowledge and sensory information from the actual environment at the time of execution. We connected PyCRAM to the \textsc{KnowRob} \ac{krr} engine to automatically infer object types, the relative poses of object features such as holes or perforated tabs, as well as affordance-related concepts such as supporting surfaces from an ontological representation of the environment. We also implemented a resolution mechanism for object poses based on \textsc{RoboSherlock}, which uses neural object detection and pose estimation models to determine the current poses of objects in the environment.

\subsubsection{Object pose estimation with RoboSherlock}
\label{sec:robosherlock}

RoboSherlock\footnote{https://robosherlock.org/publications.html} is a taskable knowledge-driven perception system based on the UIMA (Unstructured Information Management Architecture) \cite{OASIS:UIMA:2009}. RoboSherlock provides a query-answering mechanism, which acts as an interface between the perception system and the robot control program and allows the control program not only to access the world in a selective manner (i.e., faster) but also to provide prior information to RoboSherlock (e.g., the pose of a specific object given that the object is blue, cubic, ...). Given this specific query with provided prior information, RoboSherlock will plan a perception pipeline based on the capabilities and requirements of each perception expert in the knowledge base to answer the query. For instance, a 6D-pose estimation expert can provide the 3D-position and 3D-orientation of an object in the world, but requires the semantic segmentation and recognition of that object. 

 \begin{figure}[h]
 \centering
 \includegraphics[width=0.48\textwidth]{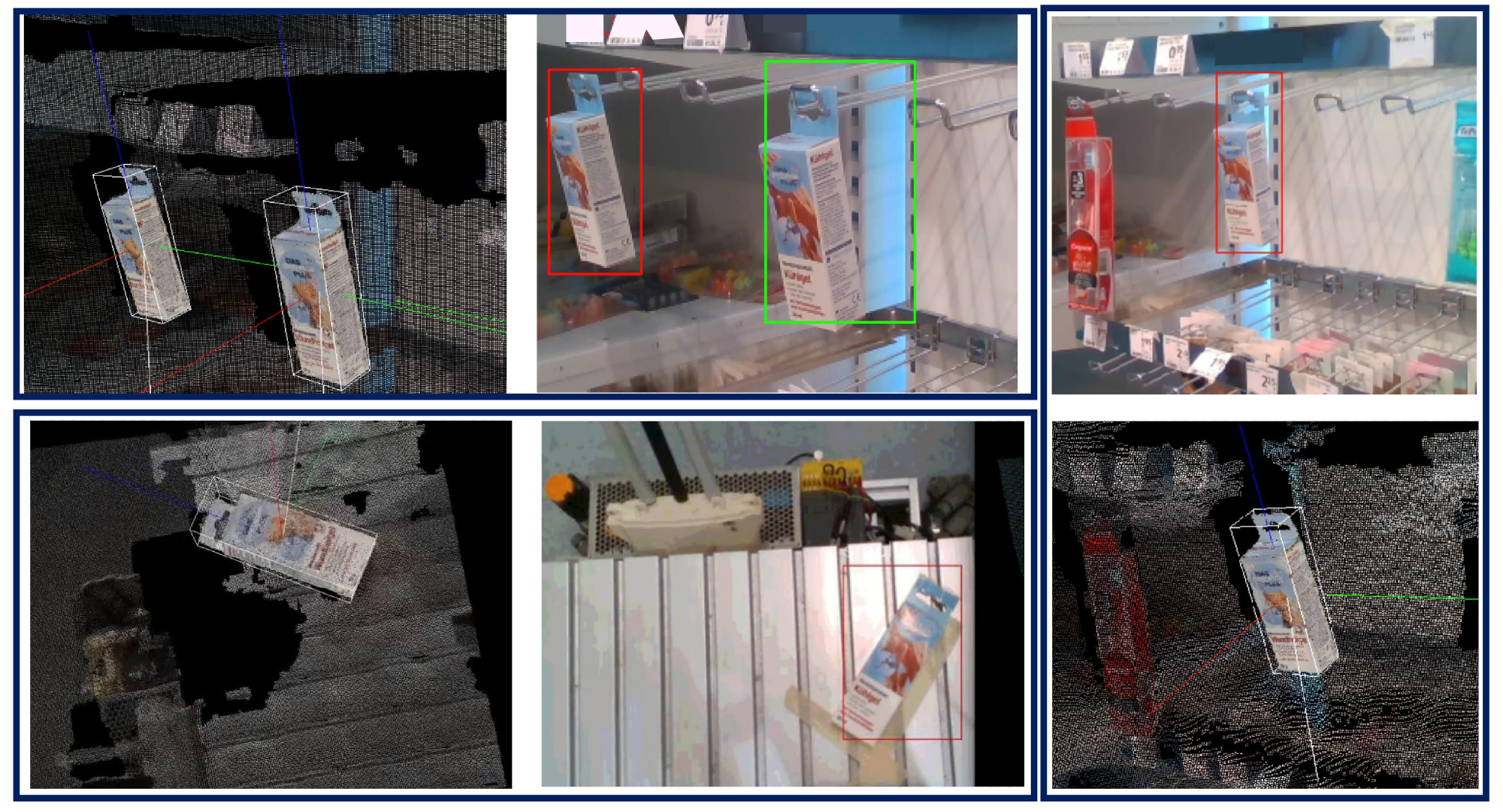}
\caption{Detection and pose estimation of objects from supermarket shelf hangers. In the three scenarios, color image illustrates the detection while point cloud image illustrates pose estimation.}
  \label{robotvqa:ilias}
\end{figure}
For this work, a core perception expert was RobotVQA (Robot Visual Question Answering) \cite{RobotVQA}, a deep learning model which is autonomously trained from physico-realistic virtual worlds (for embodied and situated training data) to infer semantic graphs of cluttered and occluded scenes. RobotVQA can be constrained to focus on a specific set of objects described in terms of their attributes (e.g., color, shape, material, category). For experiment \ref{sec:real-world-validation}, it has been constrained to detect and estimate the pose of the healing salve product depicted in Fig. \ref{fig:teaser_image}. In case of multiple similar objects in the robot's viewport, an ID resolution expert is tasked to identify and track object instances (see Fig. \ref{robotvqa:ilias}). In \cite{kazhoyan_robot_2020}, the approach has been validated in a very challenging home setting.

Given the challenges posed by object 6D pose estimation for deep learning models (e.g., symmetry, occlusions, lack of ground truth, accuracy), RobotVQA uses a knowledge-augmented approach to overcome the problem. When RobotVQA estimates attributes of objects (e.g., shape, color, category), the corresponding 3D models of these objects are retrieved from the knowledge base. Secondly, the point cloud of each object is extracted from the depth image based on the segmentation mask returned by RobotVQA. Given that this segmentation is not perfectly accurate, often the extracted point cloud of the object contains outliers, which are then removed through distance-based hierarchical clustering. Then, the ICP (Iterative Closest Point) algorithm is applied given the object's 3D model and its point cloud. Given that ICP is extremely sensitive to occlusion and initial poses, we first estimate the three main axes of the object with PCA and then launch 12 instances of ICP in parallel, each with a different annotation of the main axes (i.e., XYZ, XZY, ...). Finally, the pose returned by the ICP instance with the highest score is considered. Given that some annotations are not physically plausible in some scenarios, such as for the experiments for this work where the object on the hanger can only have two possible orientations (i.e., front-facing or back-facing), a more accurate decision can be taken.

\subsubsection{Action designator grounding}
Once all objects and locations for a given action designator have been resolved, the action designator itself can be resolved to a sequence of motion designators, atomic (possibly force-controlled) motions to be executed by the robot. \texttt{HoleOnPeg} resolves to a collision-free approach motion, a linear contact motion until contact with the peg is detected, a spiral search motion to thread the grasped object onto the peg, and a force-controlled motion to push the object onto the peg. For this work, a resolution mechanism for motion designators was implemented, which combines \textsc{KnowRob} with an industrial robot motion planner, compiler and execution environment.\footnote{ArtiMinds Robot Programming Suite, \url{https://artiminds.com}} Motion designators are resolved in three steps:
\begin{enumerate}
    \item \textit{Parameterization of motion planner and force controller:} Using the reasoning capabilities of \textsc{KnowRob}, parameters for motion planners such as target poses, maximal velocities or force controller parameters are inferred based on knowledge of the manipulated objects and the environment.
    \item \textit{Offline motion planning:} The current belief state of \textsc{KnowRob} pertaining to the state of the environment (most notably the current poses of objects and their 3D meshes, if available) are loaded into a motion planner and a collision-free motion is planned, subject to the inferred parameters and constraints.
    \item \textit{Robot code generation:} Executable, manufacturer-specific native robot code is generated for the target robot platform, including any peripheral devices (grippers, force-torque sensors).
\end{enumerate}

After resolution of all motion designators, the generated code can be executed directly on the robot.

\section{Experiments}
\label{sec:experiments}

\begin{figure*}
    \centering
    \includegraphics[width=\textwidth]{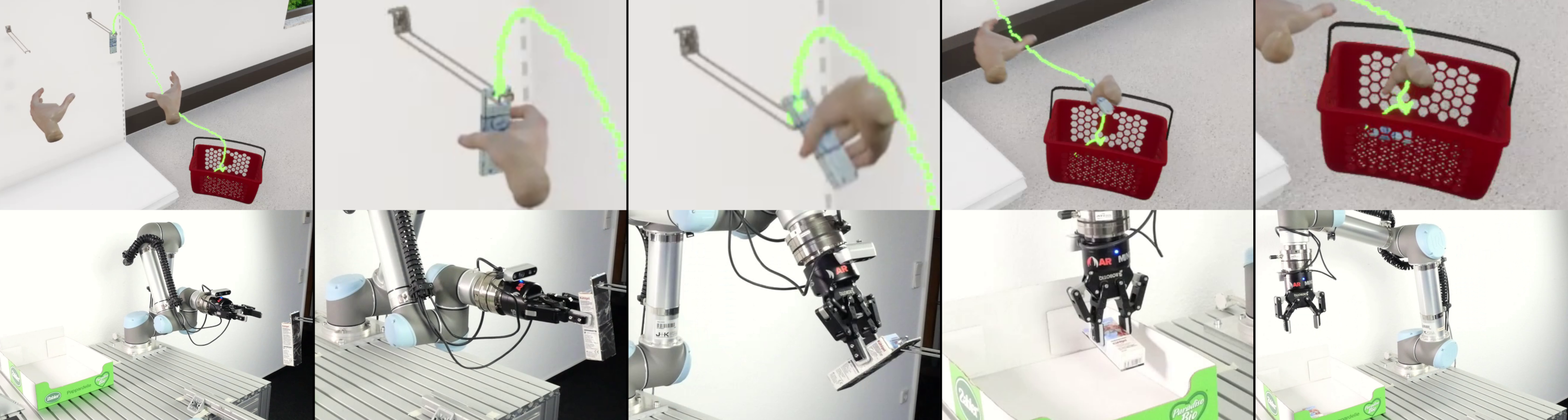}
    \caption{VR human demonstration (top) and execution of the synthesized program (bottom) for force-sensitive fetch-and-place.}
    \label{fig:img_sc1}
\end{figure*}

To assess the validity of the proposed approach for real-world robotic manipulation tasks, two application scenarios are considered. In a first scenario, the robot is tasked with taking a product down from a hanger and placing it into a shopping basket. In a second scenario, the inverse problem is considered, where the robot has to grasp a product and thread it onto a hanger. Both applications require a high degree of dexterity involving force-controlled insertion or extraction as well as sophisticated collision-free motion and grasp planning in response to changing environments.

\subsection{VR \ac{neem} Collection}

\ac{vr} demonstrations are collected using Unreal Engine 4 \cite{unrealengine} and the RobCoG\footnote{Robot Commonsense Games, \url{https://robcog.org}} framework \cite{haidu_knowrobsim_2018}, utilizing photorealistic rendering and accurate physics simulation. To evaluate the robustness of our approach with respect to variations in the demonstrations, 30 human demonstrations were recorded for each of the two application scenarios. Three box-shaped target objects, differing in width, height and depth, were used in the demonstrations to evaluate the generalization capacity of our approach across different target objects. The recorded demonstrations were stored into the \textsc{KnowRob} \ac{krr} engine as \acp{neem} (see Sec. \ref{sec:representing-demonstrations} and Fig. \ref{fig:event-timeline}), using the automatic semantic annotation pipeline presented in \cite{haidu_automated_2021}.

\subsection{Quantitative Evaluation}
In a first set of experiments, program synthesis is performed for all 60 demonstrations, the resulting program candidates are evaluated in a simulated environment and then executed on real hardware under controlled conditions. For each human demonstration, the demonstrated action sequence (\ac{neem}) is interpreted to a set of task seq
uence candidates. To that end, the Prolog reasoning engine in \textsc{KnowRob} generates all possible sequences of \texttt{Task}s consistent with the \ac{neem} and background knowledge about \texttt{Action} and \texttt{Task} semantics (see Sec. \ref{sec:interpretation-of-demonstrated-action-sequences}). This knowledge stems from TBox axioms in the \ac{soma} ontology \cite{besler_foundations_2020} and an additional Prolog rulebase containing definitions of 14 different general-purpose \texttt{Task}s, according to our proposed task semantics (see Sec. \ref{sec:representing-task-knowledge}). Due to possible ambiguities of the human demonstrations, the interpretation of an action sequence typically yields more than one candidate task sequence. The experiments chiefly test the capability of the proposed system to robustly infer executable robot programs for a wide array of human demonstrations.

Because the \ac{vr} environment differs from the real-world environment, the resulting task sequence candidates are underspecified. In this experiment, object and location designators such as the manipulated object and its location or the pose of the shopping basket are resolved by querying the knowledge base; experiment \ref{sec:real-world-validation} uses a perception system for designator resolution. For execution, a Universal Robots UR5 robot arm was used, equipped with an ATI Axia80 force-torque sensor and a Robotiq 2F-85 robotic gripper. For this set of experiments, one supermarket hanger was mounted in a fixed position (see Fig. \ref{fig:img_sc1} (bottom)). The real-world target object is a tube of healing salve in a rectangular carton box (see in Fig. \ref{fig:teaser_image}), which has been reinforced with 3D-printed plastic to withstand repeated grasping.

\subsubsection{Scenario 1: Fetching an Object From a Hanger}
\begin{figure}
    \centering
    \begin{tabular}{lccccc}\toprule
& \multicolumn{1}{c}{\small{Demonstr.}} & \multicolumn{1}{c}{\small{Interp.}} & \multicolumn{2}{c}{\small{Simulation}} & \multicolumn{1}{c}{\small{Exec.}}
\\\cmidrule(lr){2-2}\cmidrule(lr){3-3}\cmidrule(lr){4-5}\cmidrule(lr){6-6}
        & \scriptsize{\# Demos}    & \scriptsize{\# Cands.}  & \scriptsize{Plan succ.} & \scriptsize{Task succ.} & \scriptsize{Succ.} \\\midrule
O1     & 10         & 29        & 27 \tiny{(93\%)}      & 27 \tiny{(93\%)}       & 26 \tiny{(90\%)}       \\
O2     & 10         & 60        & 52 \tiny{(87\%)}     & 44 \tiny{(73\%)}     & 35 \tiny{(58\%)}      \\
O3     & 9          & 18        & 18 \tiny{(100\%)}     & 18 \tiny{(100\%)}      & 18 \tiny{(100\%)}      \\\bottomrule
\end{tabular}

    \caption{Quantitative evaluation results for force-sensitive fetch-and-place. For three different objects (O1, O2, O3), the majority of synthesized programs achieve the task in simulation and real-world execution.}
    \label{fig:results_sc1}
\end{figure}

An extract of the human demonstration and the execution of the corresponding program candidate is shown in Fig. \ref{fig:img_sc1}. Quantitative results are provided in Fig. \ref{fig:results_sc1}. For 29 out of 30 demonstrations, action interpretation produced at least one candidate task sequence, with the majority of demonstrations resulting in 2 or more candidates. In the one remaining demonstration, a glitch in the \ac{vr} engine caused the object to briefly disappear from the human avatar's hand, spawn on the ground and then re-appear in the avatar's hand, resulting in wrong semantic annotations (sudden contact with the ground plane). This prevented the reasoning engine from finding a viable task sequence for the demonstration. 
For two of the three objects, simulation of 7-13\% of the generated robot programs failed, generally due to unreachable approach, grasp or depart poses. Depending on the object, 75-100\% of the generated robot programs successfully placed the object into the basket in the simulation. For each of the 29 valid demonstrations, at least one synthesized program could be successfully executed, extracting the object from the hanger and placing it into a basket while avoiding collisions. The generated programs successful in the simulation sometimes failed in real-world execution due to exceeded force limits during extraction, or by colliding with the hanger or basket due to slight differences in the 3D models (used for collision-free planning) and the real-world objects. This can be solved by more robust parameterization of the force controller and the use of perception (e.g. 3D cameras) to ensure the planning scene matches the real-world more precisely.

\subsubsection{Scenario 2: Threading an Object Onto a Hanger}
\begin{figure}
    \centering
    \begin{tabular}{lccccc}\toprule
& \multicolumn{1}{c}{\small{Demonstr.}} & \multicolumn{1}{c}{\small{Interp.}} & \multicolumn{2}{c}{\small{Simulation}} & \multicolumn{1}{c}{\small{Exec.}}
\\\cmidrule(lr){2-2}\cmidrule(lr){3-3}\cmidrule(lr){4-5}\cmidrule(lr){6-6}
        & \scriptsize{\# Demos}    & \scriptsize{\# Cands.}  & \scriptsize{Plan succ.} & \scriptsize{Task succ.} & \scriptsize{Succ.} \\\midrule
O1     & 10         & 22        & 20 \tiny{(91\%)}      & 18 \tiny{(82\%)}       & 18 \tiny{(82\%)}       \\
O2     & 10         & 20        & 20 \tiny{(100\%)}     & 20 \tiny{(100\%)}     & 15 \tiny{(75\%)}      \\
O3     & 10         & 25        & 25 \tiny{(100\%)}     & 25 \tiny{(100\%)}      & 25 \tiny{(100\%)}      \\\bottomrule
\end{tabular}

    \caption{Quantitative evaluation results for inverse peg-in-hole. For three different objects (O1, O2, O3), the majority of synthesized programs achieve the task in simulation and real-world execution.}
    \label{fig:results_sc2}
\end{figure}

\begin{figure*}
    \centering
    \includegraphics[width=\textwidth]{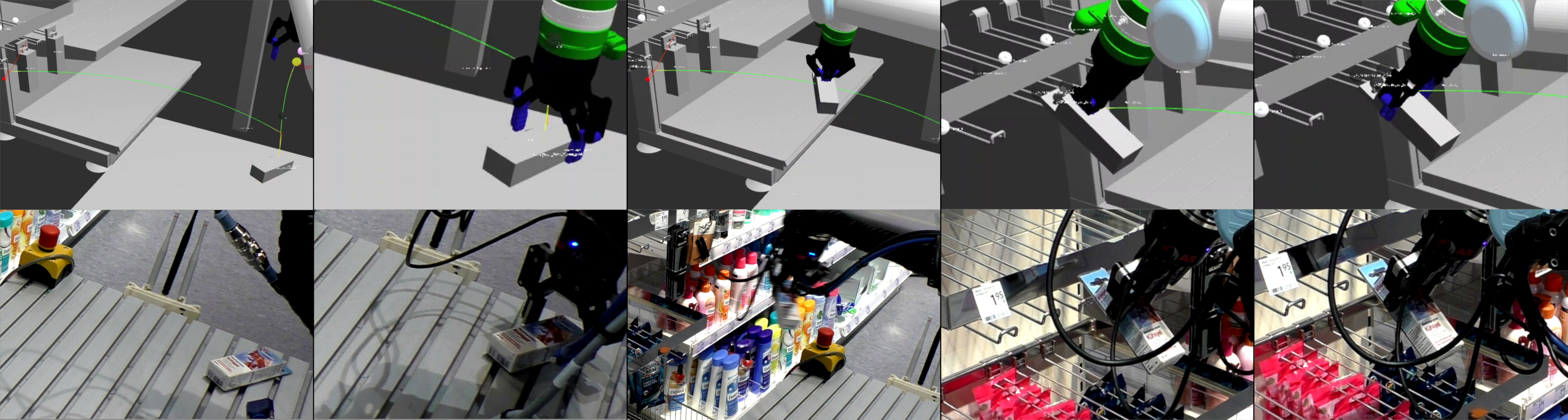}
    \caption{Simulation (top) and execution of the synthesized program (bottom) for inverse peg-in-hole in a realistic supermarket environment.}
    \label{fig:img_sc2}
\end{figure*}

A visualization of the simulation and real-world execution for scenario 2 is shown in Fig. \ref{fig:img_sc2}. Quantitative results are provided in Fig. \ref{fig:results_sc2}. For all human demonstrations, action interpretation produced at least one candidate task sequence, with the majority of demonstrations resulting in 2-4 candidates. In this scenario, all but two candidates could be simulated successfully, and all but four candidates successfully placed the object onto the hanger in the simulation. Execution was successful for 75-100\% of generated programs. The main challenge was robustly finding the tip of the hanger using force-controlled search. Integration of visual perception to precisely detect the pose of the hanger can further improve results. As for the first application scenario, at least one successfully executable robot program could be generated for each human demonstration.

\subsection{Real-World Validation}
\label{sec:real-world-validation}
A second set of experiments aims at validating the proposed approach in a realistic supermarket environment. To that end, we realized both application scenarios in a laboratory for retail robotics \cite{costanzo_manipulation_2020}, which accurately models the furniture, products and lighting conditions in a real supermarket. For execution, Donbot\footnote{https://ai.uni-bremen.de/research/robots/donbot} is used, which features a mobile platform and a Universal Robots UR5 collaborative robot arm. For this experiment, an ATI Axia80 force-torque sensor, a Robotiq 2F-85 robotic gripper and a flange-mounted Intel RealSense camera have been added.
Both scenarios were solved successfully in this more realistic setting (see Fig. \ref{fig:img_sc2} for a visualization of the experiments for Scenario 2). RoboSherlock (see Sec. \ref{sec:robosherlock}) was used to ensure that changes in the pose of the manipulated objects were reflected in the knowledge base. Our framework generated successful robot control programs for each task given a single human demonstration, even when the pose of the manipulated object was changed.

\section{Related Work}
\label{sec:related_work}
Most recent work in program synthesis is dedicated to one of two central challenges: \textit{Action (or task) recognition}, the process of identifying the intention behind e.g. a human demonstration or a natural-language instruction, and \textit{task execution}, the process of generating robust low-level robot control commands to solve high-level tasks in real environments. Several action recognition frameworks based on deep neural networks have been proposed \cite{sun_human_2022,kong_human_2022}. However, for the purpose of program synthesis, it is highly advantageous to detect not only the human activity itself, but also to extract rich semantic information about the agent, objects and execution context. To that end, knowledge-based approaches to human activity recognition have been proposed \cite{ramirez-amaro_transferring_2017,haidu_action_2016,bates_-line_2017}. \acp{neem} as a universal exchange format for semantically rich, annotated human or robot experience data have been proposed in \cite{beetz_know_2018}. In \cite{haidu_automated_2021}, Haidu et al. propose a mechanism to automatically parse human demonstrations to \acp{neem}. The present work is the first to leverage \acp{neem} as an input modality for program synthesis.

Task execution has been traditionally addressed by mapping tasks to robot skills in a skill library \cite{ramirez-amaro_transferring_2017,ramirez-amaro_understanding_2015}. Alternatively, \ac{tamp} approaches have been proposed, which derive a set of constraints from a task description and solve a multi-horizon planning problem \cite{kaelbling_hierarchical_2011,schmitt_planning_2019,diehl_optimizing_2021}. Underspecified plans \cite{beetz_cram_2010} and plan specialization \cite{koralewski_self-specialization_2019} suggest an alternative approach, whereby high-level plans are incrementally specialized to executable programs by way of reasoning and perception. In prior work \cite{kazhoyan_towards_2020,kazhoyan_robot_2020}, underspecified plans were written manually upfront. This work automatically generates underspecified plans directly from human demonstrations.

Recent work has increasingly leveraged deep learning for program synthesis in the domain of robotics \cite{xu_neural_2018,sun_neural_2018}. \acp{llm} have been used to this end with particular success \cite{liang_code_2023,singh_progprompt_2022}. However, in complex long-horizon force-controlled manipulation scenarios, purely neural program synthesis approaches struggle to address the dual requirement of generating policies which generalize well, but at the same time solve a given task with high precision. The knowledge-based action interpretation and task execution mechanisms proposed in this work provide cognitive mechanisms which address these challenges.

\section{Conclusion}

This work proposes and describes a knowledge-driven approach to robot program synthesis for real-world open-ended manipulation tasks. Embedded into a state-of-the-art cognitive architecture, it leverages sophisticated knowledge representations and reasoning algorithms to interpret the task-level intentions of human demonstrations in \ac{vr}, generate a generalized motion plan and transform it into executable robot code via reasoning, path planning and knowledge-enabled perception.

\subsection{Discussion and Future Work}

Designing a program synthesis system around a common knowledge representation and using a shared reasoning engine permits a high degree of generalization and integration. The \ac{soma} foundational ontology, \acp{neem} as an exchange format for experience data and integration in the \textsc{Cram} cognitive architecture allow for sharing knowledge and reasoners between components. Moreover, ontology-based knowledge representation allows our system to be highly generalizable, as task semantics can be specified at a very general level (see Sec. \ref{sec:representing-task-knowledge}): In our experiment, the same task knowledge could be used to infer robot control programs for both fetch-and-place and peg-in-hole tasks. The same mechanisms permit generalization to domains beyond supermarkets, such as industrial or household settings.

One limitation of the presented approach is that the range of tasks that can be solved is limited to the contents of the knowledge base, and that extending the knowledge base requires manual adding of task knowledge (e.g. pre-, runtime- and postconditions). Moreover, the approach implies a trade-off between filtering feasible candidate programs in simulation (which risks rejecting good candidates if the simulation differs from reality), and trying out the remaining candidates in the real world (which requires time and resources). These shortcomings can be addressed by integrating learning approaches such as interactive task learning \cite{gluck_interactive_2019}, which will be the subject of further investigation.

\section*{Acknowledgments}

This work has been partly funded by the German ministry of education and research (BMBF) as part of the ILIAS project (reference no. 01DR19001B).

\bibliographystyle{kr}
\bibliography{bibfile, bibfile_franklin}

\end{document}